\pdfoutput=1
 
\DeclareFontSeriesDefault[rm]{bf}{sbc}

\documentclass[twoside,11pt]{article}

\usepackage{amsmath}
\usepackage{blindtext}
\usepackage[table]{xcolor}
\usepackage[preprint]{jmlr2e}
\usepackage{beton}

\usepackage{lastpage}

\ShortHeadings{Extending F1 metric, probabilistic approach}{Extending F1 metric, probabilistic approach}
\firstpageno{1}

\begin{document}

\title{Extending F1 metric, probabilistic approach}

\author{\name Mikołaj Sitarz \email sitarz@refaba.com \\
  \includegraphics[width=3em]{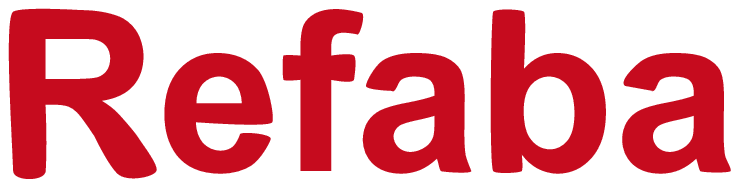} \\
  2022 Kraków \\
}

\maketitle

\begin{abstract}
   This article explores the extension of well-known $\mathrm{F}_1$ score used for assessing the performance of binary classifiers.  We
   propose the new metric using probabilistic interpretation of precision, recall, specificity, and negative predictive value. We describe
   its properties and compare it to common metrics.  Then we demonstrate its behavior in edge cases of the confusion
   matrix. Finally, the properties of the metric are tested on binary classifier trained on the real dataset.
\end{abstract}


\section{Background}

The $\mathrm{F}_1$ metric -- as described by \cite{sasaki:07} -- is commonly used to evaluate the performance of binary machine learning
classifiers.  Calculated as a harmonic mean of precision and recall (see section \ref{metrics}) gains an advantage over less complex metrics
like accuracy. Especially when used against imbalanced datasets.  The properties of $\mathrm{F}_1$ and methods of maximizing the expected metric
value, were precisely described, and analyzed from the theoretical and experimental point of view by \cite{lipton:2014}.

$\mathrm{F}_1$ is often criticized as an evaluation metric. The main axis of that critique is lack of the dependency on \emph{true negatives} -
pointed among the others by \cite{powers:2010} and \cite{hand2018note}. Another of its drawbacks is asymmetry -- it may give different score
when the dataset labeling is changed (positives labeled as negatives and negatives labeled as positives).
These facts make it unreliable as a metric in certain cases.

As a cure for these $\mathrm{F}_1$ problems -- $\mathrm{MCC}$ is often pointed to -- like presented by \\ \cite{chicco:2020:mcc:f1}. On the other hand
sometimes, researchers prefer to use them both ``cooperating'' -- like \cite{cao:2020:mcc}. To this last aspect, we will return later in our article.

\section{Common metrics} \label{metrics}
\subsection{Basic and composite metrics}

Let us first list the basic building blocks of which the binary classifier metrics are composed:
\begin{itemize}
\item $\mathrm{TP}$ - true positives (positive samples classified as positive), 
\item $\mathrm{FN}$ - false negatives (positive samples misclassified as negative),
\item $\mathrm{TN}$ - true negatives (negative samples classified as negative),
\item $\mathrm{FP}$ - false positives (negative samples misclassified as positive),
\end{itemize}

Based on them, the \emph{precision} and \emph{recall} (also called \emph{sensitivity}) metrics were defined:
\begin{displaymath}
  \mathrm{PREC} = \frac{\mathrm{TP}}{\mathrm{TP}+\mathrm{FP}}
\end{displaymath}

\begin{displaymath}
    \mathrm{REC} = \frac{\mathrm{TP}}{\mathrm{TP} + \mathrm{FN}}
\end{displaymath}

Then $\mathrm{F}_1$ is defined as a harmonic mean of two above:
\begin{displaymath}
  \mathrm{F}_1 = \frac{2}{\frac{1}{\mathrm{PREC}} + \frac{1}{\mathrm{REC}}} = \frac{\mathrm{TP}}{\mathrm{TP} + \frac{1}{2} (\mathrm{FP} + \mathrm{FN})}
\end{displaymath}

We will also use another metric called \emph{specificity}:
\begin{displaymath}
  \mathrm{SPEC} = \frac{\mathrm{TN}}{\mathrm{TN} + \mathrm{FP}}
\end{displaymath}

\emph{Precision} and \emph{recall} are more popular in machine learning publications, while the medical ones usually prefer \emph{recall} and \emph{specificity}.
The latter two are the base components of \emph{Youden index} (also known as \emph{informedness}) defined by \cite{youden1950index}:
\begin{displaymath}
  \mathrm{J} = \mathrm{REC} + \mathrm{SPEC} - 1
\end{displaymath}

We also want to include another metric called - \emph{negative predictive value} - $\mathrm{NPV}$ in this overview:
\begin{displaymath}
  \mathrm{NPV} = \frac{\mathrm{TN}}{\mathrm{TN}+\mathrm{FN}}
\end{displaymath}

Finally, $\mathrm{NPV}$ is a component of \emph{markedness} (see: \cite{powers:2010}):
\begin{displaymath}
  \mathrm{MK} = \mathrm{PREC} + \mathrm{NPV} - 1
\end{displaymath}
The presented list obviously is not comprehensive - it does not exhaust all the metrics in use. However, we will focus on them in our
discussion.

Summarizing: in the further sections we will be considering the following plain metrics: $\mathrm{PREC}$, $\mathrm{REC}$, $\mathrm{SPEC}$,
$\mathrm{NPV}$, and following composite metrics: $\mathrm{F}_1$, $\mathrm{J}$, $\mathrm{MK}$. 

\subsection{Matthews correlation coefficient}

On a separate note, the $\mathrm{MCC}$ metric deserves attention. \emph{Matthews correlation coefficient} known under several other forms, in its present form defined by \cite{matthews:1975}.
$\mathrm{MCC}$ is a Pearson correlation coefficient calculated for two binary sequences: the original sample values (positives and negatives) and the values returned by classifier. In terms of
$\mathrm{TP}$, $\mathrm{FN}$, $\mathrm{TN}$, $\mathrm{FP}$ values it can be calculated as follows:

\begin{displaymath}
  \mathrm{MCC} = \frac{\mathrm{TP} \cdot \mathrm{TN} - \mathrm{FP} \cdot \mathrm{FN}}{\sqrt{(\mathrm{TP} + \mathrm{FP}) (\mathrm{TP} + \mathrm{FN}) (\mathrm{TN} + \mathrm{FP}) (\mathrm{TN} + \mathrm{FN})}}
\end{displaymath}

\section{Probabilistic approach -- focusing on conditional probabilities} \label{probabilistic}

The probabilistic interpretation of $\mathrm{F}_1$, $\mathrm{PREC}$ and $\mathrm{REC}$ has been comprehensively presented by \cite{goutte2005probabilistic}.
We want to attack the problem from the opposite side and start with the definition of 4 conditional probabilities which values we want to maximize when designing a binary classifier:
\begin{itemize}
\item $P(+ \mid C{+})$ -- the probability that the sample is positive, provided the classifier result was positive.
\item $P(C{+} \mid +)$ -- the probability that the classifier result will be positive, provided the sample is positive.
\item $P(C{-} \mid -)$ -- the probability that the classifier result will be negative, provided the sample is negative.
\item $P(- \mid C{-})$ -- the probability that the sample is negative, provided the classifier result was negative.
\end{itemize}
Given all these conditional probabilities, we can require a valid classifier to produce results for which each is close to 1.
Basing on this requirement, we are building the new metric $\mathrm{P}_4$, demanding it to have the following properties:
\begin{enumerate}
\item The metric value is limited to the given range: $\mathrm{P}_4 \in [0, 1]$.
\item When any of the four conditional probabilities tends to zero, $\mathrm{P}_4$ metric also tends to zero regardless of the values of the other probabilities.
\item When all the four conditional probabilities tend to one, $\mathrm{P}_4$ metric also tends to one.
\end{enumerate}

Now let us quantitatively describe each of these probabilities:
\begin{equation}\label{p_prec}
  P(+ \mid C{+}) =  \frac{\mathrm{TP}}{\mathrm{TP}+\mathrm{FP}} = \mathrm{PREC}
\end{equation}

\begin{equation}\label{p_rec}
  P(C{+} \mid +) = \frac{\mathrm{TP}}{\mathrm{TP} + \mathrm{FN}} = \mathrm{REC}
\end{equation}

\begin{equation}\label{p_spec}
  P(C{-} \mid -) =  \frac{\mathrm{TN}}{\mathrm{TN} + \mathrm{FP}} = \mathrm{SPEC}
\end{equation}

\begin{equation}\label{p_npv}
  P(- \mid C{-}) =  \frac{\mathrm{TN}}{\mathrm{TN} + \mathrm{FN}} = \mathrm{NPV}
\end{equation}

Coming back to the composite metrics mentioned before: $\mathrm{F}_1$ captures only probabilities \eqref{p_prec} and \eqref{p_rec}, $\mathrm{J}$ is
based on \eqref{p_rec} and \eqref{p_spec}, while $\mathrm{MK}$ depends on \eqref{p_prec} and \eqref{p_npv}.
We have not yet met a metric that refers directly to all four probabilities at once.  This fact imposes a certain desire to combine all of
them into a single measure. So, let us finally define $\mathrm{P}_4$ as a harmonic mean of all the four conditional probabilities:

\begin{displaymath}
  \mathrm{P}_4 = \frac{4}{\frac{1}{\mathrm{PREC}} + \frac{1}{\mathrm{REC}} + \frac{1}{\mathrm{SPEC}} + \frac{1}{\mathrm{NPV}}}
\end{displaymath}
Thus, we get:
\begin{displaymath}
  \mathrm{P}_4 = \frac{4\cdot\mathrm{TP}\cdot\mathrm{TN}}{4\cdot\mathrm{TP}\cdot\mathrm{TN} + (\mathrm{TP} + \mathrm{TN}) \cdot (\mathrm{FP} + \mathrm{FN})}
\end{displaymath}
The newly defined $\mathrm{P}_4$ metric satisfies all the three requirements we defined above. This is due to the properties of the harmonic mean.
What is more, for the requirements 2 and 3 the inverse implication is also true:
\begin{enumerate}
  \setcounter{enumi}{3} 
\item When $\mathrm{P}_4$ metric tends to zero, at least one of the conditional probabilities is close to zero.
\item When $\mathrm{P}_4$ metric tends to one, all the probabilities are close to one.
\end{enumerate}
$\mathrm{P}_4$ is also symmetrical with respect to dataset labels swapping (similarly to the $\mathrm{MCC}$ metric), as opposed to $\mathrm{F}_1$ -- see appendix \ref{symmetry}.
In the coming sections, we will take a closer look at the newly defined metric and compare its properties with those of the commonly known metrics.

\section{Edge cases}
\subsection{Confusion matrix}
It is essential that when analyzing the performance of a classifier, it should not be considered in isolation from the population to which it applies.
The same classifier used on two populations having different sample distributions -- will lead to the different performance metric values.
Therefore, a convenient way to present the performance of a classifier with respect to a given population is a confusion matrix:

\begin{center}
  \begin{tabular}{c | c | c}
    & Actual positive & Actual negative \\
    & TP+FN & FP+TN \\
    \hline
    Classified positive &  \cellcolor{green!20} &  \cellcolor{red!20} \\
    TP + FP & \cellcolor{green!20} TP & \cellcolor{red!20} FP \\
     &  \cellcolor{green!20} &  \cellcolor{red!20} \\
    \hline
    Classified negative &  \cellcolor{red!20} &  \cellcolor{green!20} \\
    FN + TN & \cellcolor{red!20} FN &  \cellcolor{green!20} TN \\
    &  \cellcolor{red!20} &  \cellcolor{green!20} \\
  \end{tabular}
\end{center}

We will use the shorten version of confusion matrix, in our demonstration:

\begin{displaymath}
  \mathbf{C} = \begin{bmatrix}
    \mathrm{TP} & \mathrm{FP} \\
    \mathrm{FN} & \mathrm{TN} \\
  \end{bmatrix}
\end{displaymath}

To show the properties of $\mathrm{P}_4$ against other metrics, we will present four examples of confusion matrices. For
each of them, one of the conditional probabilities \eqref{p_prec}, \eqref{p_rec}, \eqref{p_spec}, \eqref{p_npv} is close to $0$, while the others are moderately close to $1$.
In each of the cases presented, we use a simulated classifier and a population of $10000$ samples.

Because some of the metrics presented above, are ranged in $[-1, 1]$ than in $[0, 1]$, we must scale them first, to compare with the other ones.
Thus, we will be using:
\begin{displaymath}
  \mathrm{MCC}' = (\mathrm{MCC} + 1) / 2
\end{displaymath}
\begin{displaymath}
  \mathrm{J}' = (\mathrm{J} + 1) / 2
\end{displaymath}
\begin{displaymath}
  \mathrm{MK}' = (\mathrm{MK} + 1) / 2
\end{displaymath}

\subsection{Case 1 - ``alarming precision''}

This is a classic case in which $\mathrm{F}_1$ shines. The population is highly imbalanced in favor of negative samples.
The classifier's performance on positive samples is $90\%$, the same on negative ones. Thus, we have the following confusion matrix:

\begin{displaymath}
  \mathbf{C_1} = \begin{bmatrix}
    45 & 995 \\
    5 & 8955 \\
  \end{bmatrix}
\end{displaymath}

In this case, the four conditional probabilities are as follows:

\begin{center}
  \begin{tabular}{c | l}
    Conditional Probability & Value \\
    \hline
    $P(+ \mid C{+})$ & 0.0433 \cellcolor{yellow!20}\\
    $P(C{+} \mid +)$ & 0.9000\\
    $P(C{-} \mid -)$ & 0.9000\\
    $P(- \mid C{-})$ & 0.9994\\
  \end{tabular}
\end{center}

And now let us look at how these affect the values of our metrics:

\begin{center}
  \begin{tabular}{c | l}
    Metric & Value \\
    \hline
    $\mathrm{P}_4$ & 0.1519 \cellcolor{yellow!20}\\
    $\mathrm{F}_1$ & 0.0826 \cellcolor{yellow!20}\\
    $\mathrm{MCC}'$ & 0.5924 \\
    $\mathrm{J}'$ & 0.9000 \cellcolor{red!10}\\
    $\mathrm{MK}'$ & 0.5214 \\
  \end{tabular}
\end{center}

We can identify three groups of metrics in the table above:
\begin{itemize}
\item ``Close to zero'' (yellow) group -- having two members:  $\mathrm{P}_4$ and $\mathrm{F}_1$.
\item ``Middle of the range'' (white) group -- $\mathrm{MCC}'$ and $\mathrm{MK}'$ -- still reacting OK on the situation.
\item ``Ignoring'' (red) group -- metric $\mathrm{J}'$ -- no proper reaction on the low conditional probability: $P(+ \mid C{+})$.
\end{itemize}

\subsection{Case 2 - ``alarming negative predictive value''}

This case can be simply obtained from ``Case 1'' by re-labeling the samples - naming  the ``positives'' as ``negatives'' and vice versa. Thus, we have the following confusion matrix:

\begin{displaymath}
  \mathbf{C_2} = \begin{bmatrix}
    8955 & 5 \\
    995 & 45 \\
  \end{bmatrix}
\end{displaymath}

Probabilities:

\begin{center}
  \begin{tabular}{c | l}
    Conditional Probability & Value \\
    \hline
    $P(+ \mid C{+})$ & 0.9994\\
    $P(C{+} \mid +)$ & 0.9000\\
    $P(C{-} \mid -)$ & 0.9000\\
    $P(- \mid C{-})$ & 0.0433 \cellcolor{yellow!20}\\
  \end{tabular}
\end{center}

That gives the following metric values:

\begin{center}
  \begin{tabular}{c | l}
    Metric & Value \\
    \hline
    $\mathrm{P}_4$ & 0.1519 \cellcolor{yellow!20}\\
    $\mathrm{F}_1$ & 0.9471 \cellcolor{red!20}\\
    $\mathrm{MCC}'$ & 0.5924 \\
    $\mathrm{J}'$ & 0.9000 \cellcolor{red!20}\\
    $\mathrm{MK}'$ & 0.5214 \\
  \end{tabular}
\end{center}

As we can see, $\mathrm{F}_1$ is the only metric that changed its value after the label swap. This clearly shows the problem with its
asymmetry.  Contrasting to the previous case, $\mathrm{F}_1$ is completely not noticing one of the key probabilities being close to zero.
The other metrics considered have not changed compared to the ``Case 1''.

\subsection{Case 3 - ``alarming recall''}

This case represents a typical situation when the classifier is over-predicting in favor of negative results, resulting a particularly superior
performance on the negative samples and deficient performance on the positive samples. The population contains 10\% positive samples.
So, there we have the confusion matrix:

\begin{displaymath}
  \mathbf{C_3} = \begin{bmatrix}
    50 & 9 \\
    950 & 8991 \\
  \end{bmatrix}
\end{displaymath}

Conditional probabilities:

\begin{center}
  \begin{tabular}{c | l}
    Conditional Probability & Value \\
    \hline
    $P(+ \mid C{+})$ & 0.8475\\
    $P(C{+} \mid +)$ & 0.0500 \cellcolor{yellow!20}\\
    $P(C{-} \mid -)$ & 0.9990\\
    $P(- \mid C{-})$ & 0.9044\\
  \end{tabular}
\end{center}

Values of the metrics:

\begin{center}
  \begin{tabular}{c | l}
    Metric & Value \\
    \hline
    $\mathrm{P}_4$ & 0.1718 \cellcolor{yellow!20}\\
    $\mathrm{F}_1$ & 0.0944 \cellcolor{yellow!20}\\
    $\mathrm{MCC}'$ & 0.5960 \\
    $\mathrm{J}'$ & 0.5245 \\
    $\mathrm{MK}'$ & 0.8759 \cellcolor{red!20}\\
  \end{tabular}
\end{center}

As we see $\mathrm{F}_1$ is back in the league. $\mathrm{MCC}'$ and $\mathrm{J}'$ playing well.
And a red card is given to a player of the visiting team: $\mathrm{MK}'$.

\subsection{Case 4 - ``alarming specificity''}
The last case represents the inversion of ``Case 3''. The classifier over-predicts in favor of positive results -- having particularly superior performance on positive results
and deficient performance on the negative ones. The population contains 10\% negative samples. So, let us look at the confusion matrix, probabilities, and the metrics:

\begin{displaymath}
  \mathbf{C_4} = \begin{bmatrix}
    8991 & 950 \\
    9 & 50 \\
  \end{bmatrix}
\end{displaymath}

\begin{center}
  \begin{tabular}{c | l}
    Conditional Probability & Value \\
    \hline
    $P(+ \mid C{+})$ & 0.9044\\
    $P(C{+} \mid +)$ & 0.9990\\
    $P(C{-} \mid -)$ & 0.0500 \cellcolor{yellow!20}\\
    $P(- \mid C{-})$ & 0.8475\\
  \end{tabular}
\end{center}

\begin{center}
  \begin{tabular}{c | l}
    Metric & Value \\
    \hline
    $\mathrm{P}_4$ & 0.1718 \cellcolor{yellow!20}\\
    $\mathrm{F}_1$ & 0.9494 \cellcolor{red!20}\\
    $\mathrm{MCC}'$ & 0.5960 \\
    $\mathrm{J}'$ & 0.5245 \\
    $\mathrm{MK}'$ & 0.8759 \cellcolor{red!20}\\
  \end{tabular}
\end{center}

$\mathrm{MK}'$ again occupies the ``red'' group, this time together with $\mathrm{F}_1$ as a companion. $\mathrm{P}_4$ obviously -- works as designed.

\subsection{Summary}
As we have seen in the four edge cases: none of the considered, existing so far compound metrics ($\mathrm{F}_1$, $\mathrm{J}$, $\mathrm{MK}$),
guarantees correct behavior in all of them.
$\mathrm{MCC}$ as a correlation coefficient here represents a separate category and stands out positively against its background.
And even though it does not reach values near its minimum in edge cases, its performance should still be considered satisfactory.

What is not surprising, however, is the behavior of the newly defined $\mathrm{P}_4$ metric itself -- it reaches a correspondingly low value
every time, and this is due to the very assumptions on which it was based.

\section{$\mathrm{P}_4$ compared to other metrics}
In the following subsections we will again compare $\mathrm{P}_4$ against four traditional metrics: $\mathrm{MCC}$,  $\mathrm{F}_1$, $\mathrm{J}$ and $\mathrm{MK}$.
This time, covering quasi-continuous range of cases. As in previous cases, we use a simulated classifier and a fixed population size of $10000$.

\subsection{Metrics vs population balance}
In the experiment we fixed the following parameters:
\begin{itemize}
\item The ratio of \emph{true positives} to the \emph{actual positives} ($\mathrm{TPR}$ -- \emph{true positive rate}) is fixed and equals $0.1$.
\item The ratio of \emph{true negatives} to the \emph{actual negatives} ($\mathrm{TNR}$ -- \emph{true negative rate}) is fixed and equals $0.1$.
\end{itemize}
Thus, using our rather poor ``classifier'', we are observing how the values of each metric change as a function of: \emph{actual positives} to population size ratio.
The result can be seen on the charts below:

\begin{center}
  \includegraphics[width=0.8\textwidth]{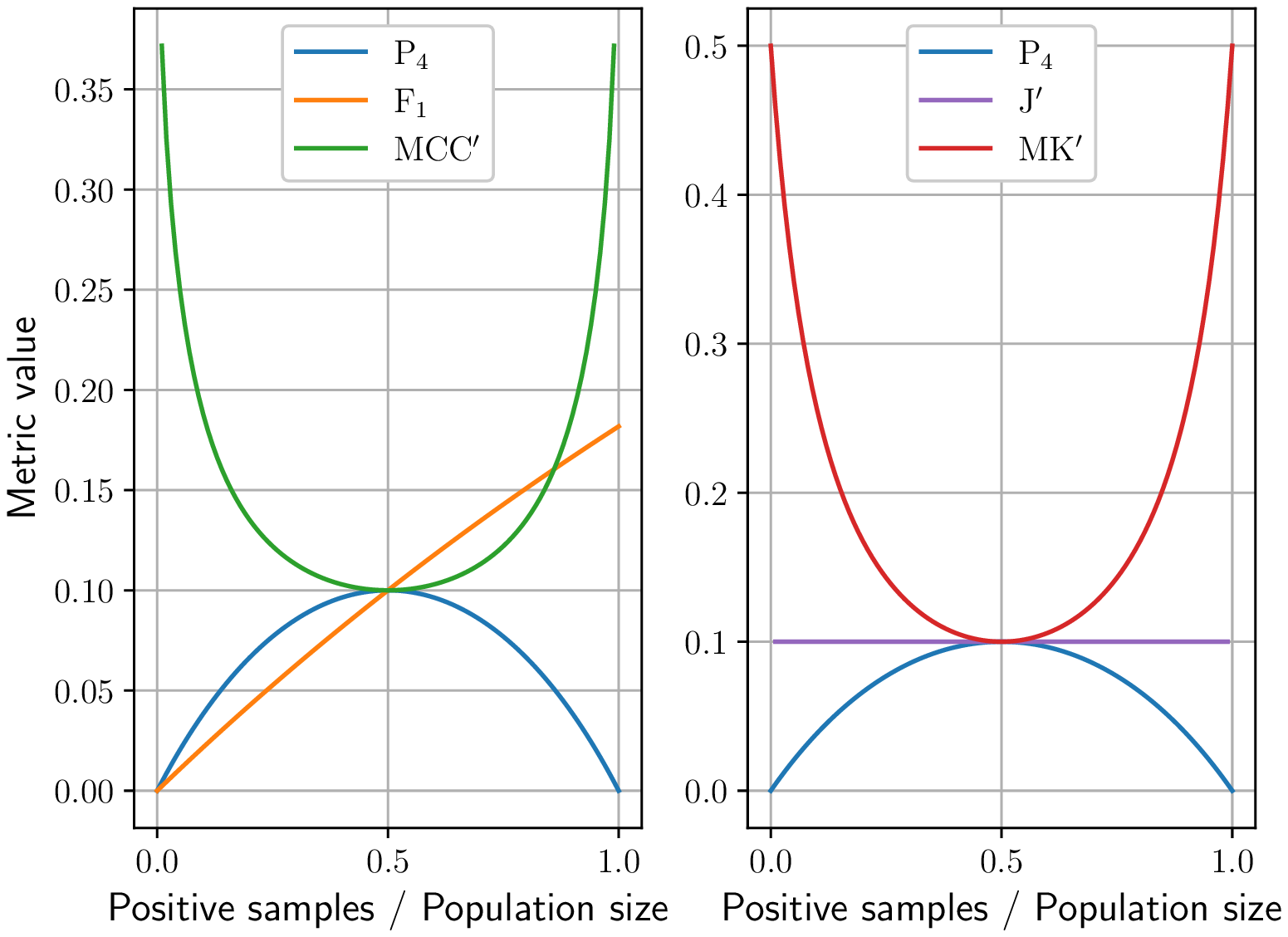}
\end{center}

On the left chart, attention is drawn to the symmetrical shape determined by $\mathrm{MCC'}$ and $\mathrm{P}_4$ curves, while $\mathrm{F}_1$
follows its own path. On the right chart we have a similar symmetrical shape as in the previous one, but this time $\mathrm{P}_4$ is
accompanied by $\mathrm{MK'}$. Youden index is not sensitive to the population balance change. The example presented here is distant
from the results obtained from the classifiers encountered on a daily basis but has the advantage of capturing differences between the metrics studied.

\subsection{Metrics vs true positive rate}
Let us see a bit more realistic example now. The following parameters are fixed now:
\begin{itemize}
\item The ratio of \emph{true negatives} to the \emph{actual negatives} is fixed and equals $0.8$.
\item The ratio of \emph{actual positives} to the population size is also fixed and equals $0.95$.
\end{itemize}
Then we are observing how the metric changes as a function of $\mathrm{TPR}$ (\emph{true positive rate}).

\begin{center}
  \includegraphics[width=0.8\textwidth]{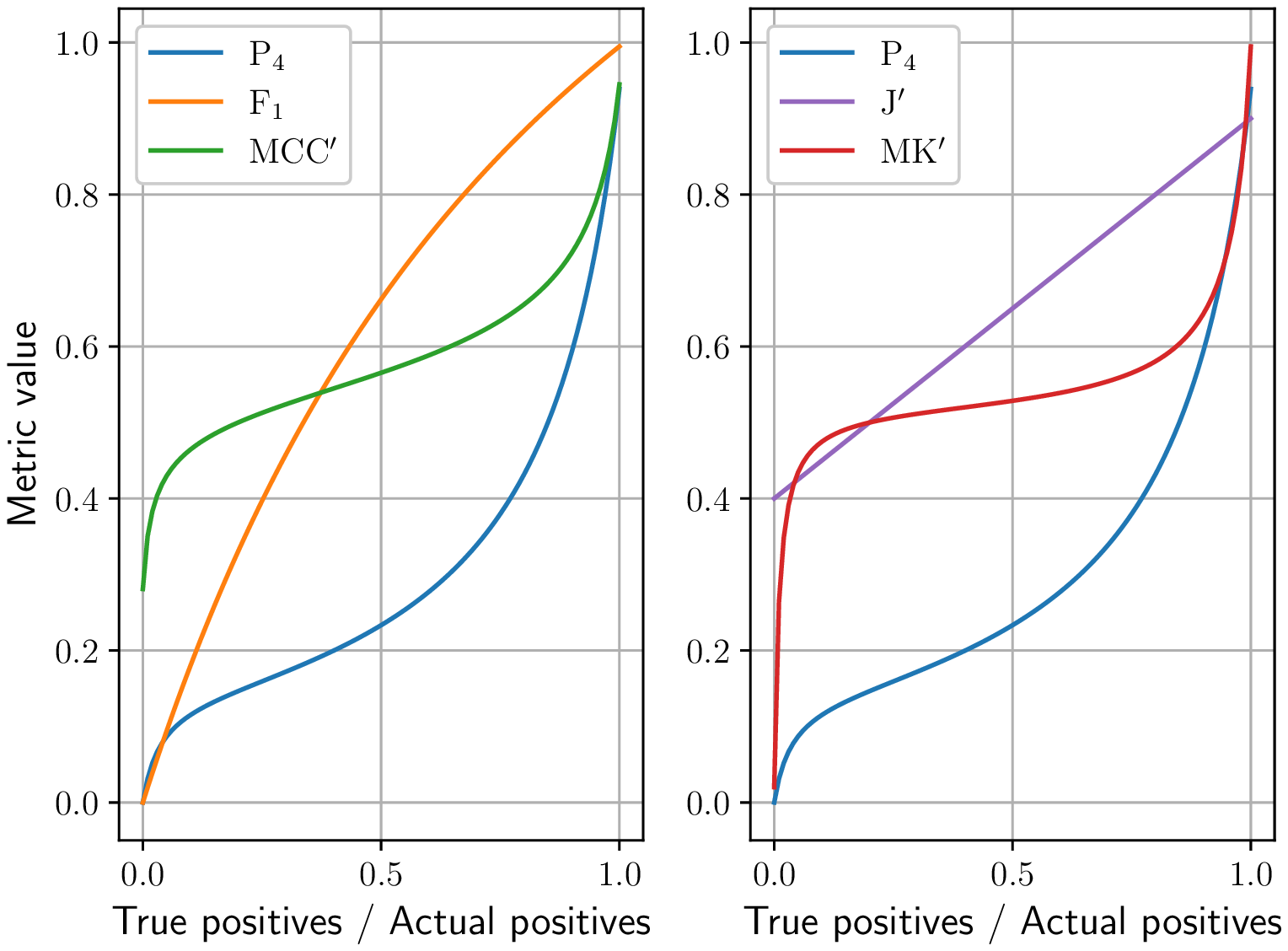}
\end{center}
In this case, the course of $\mathrm{P}_4$ and $\mathrm{MCC'}$ metrics is similar in nature, however for the first half of $\mathrm{TPR}$'s range the difference between them is roughly $\approx 0.3$.
They reach full agreement at the end of the plot. At this point the difference between them and $\mathrm{F}_1$ equals 0.05 -- which is \emph{actual negatives} to the population size ratio.

\subsection{Summary}
The charts presented here do not exhaust all the relationships between the metrics being discussed.  Many of the aspects are left
undiscussed due to the short form of this paper. We are also not analyzing the base building blocks of the composite metrics:
$\mathrm{PREC}$, $\mathrm{REC}$, $\mathrm{SPEC}$, $\mathrm{NPV}$, because the results they give are simpler than the presented ones. The
analysis of \emph{accuracy} is skipped because of the same reason. The following conclusions can be drawn from the charts:
\begin{itemize}
\item $\mathrm{P}_4$ and $\mathrm{MCC}$ behave differently in extreme case, however they tend to behave similarly in the more real-world case.
\item $\mathrm{F}_1$ is oversimplified compared with $\mathrm{P}_4$ and $\mathrm{MCC}$. 
\item $\mathrm{J'}$ behaves linear in presented cases
\end{itemize}

\section{$\mathrm{P}_4$ metric in use}
To demonstrate the properties and usefulness of $\mathrm{P}_4$, we will check how it behaves on a real dataset.
To achieve this goal the technique derived from the ``Receiver operating characteristic'' method will be used.

We will use well-known Breast Cancer Wisconsin dataset provided by UCI Machine Learning Repository (http://archive.ics.uci.edu/ml), with a
help of Scikit-Learn package -- \cite{scikit-learn}. It's a set consisting of 569 samples, 30 dimensions. The samples contain various
characteristics of biological cell nuclei (radius, texture, symmetry etc.) and the cancer binary classification: malignant/benign.

\subsection{Receiver operating characteristic}
``Receiver operating characteristic'' (ROC) - is commonly used technique for assessing the trade-off between \emph{recall} and
\emph{specificity}.  It's used together with the classifiers that, as a result -- give a probability of being positive for the sample --
like for example logistic regression. To obtain an answer ``positive''/''negative'', we must decide on a specific
probability threshold $\tau$ above which we consider the sample positive. Using ROC technique -- we are starting from $\tau_0=0$ and iterating,
increasing it by $\Delta \tau$ until $\tau_n=1$ is reached. For each $\tau_i$ we calculate confusion matrix $\mathbf{C_i}$ and thus the
\emph{precision}-\emph{recall} pair. Then we plot the curve on the $\mathrm{REC}$ vs $\mathrm{SPEC}$ chart. That plot gives us an
insight into the characteristic of the classifier-dataset pair and allows choosing optimal $\tau$ threshold.

This method has been creatively adapted by \cite{cao:2020:mcc}. Instead of \emph{precision}-\emph{recall} pair,
$\mathrm{MCC}$-$\mathrm{F}_1$ pair has been used in their case, allowing more unambiguous result and easier selection of the optimum.

\subsection{$\mathrm{MCC}$-$\mathrm{F}_1$ and $\mathrm{MCC}$-$\mathrm{P}_4$  curves}
We will use the same technique as mentioned above but also including the $\mathrm{P}_4$ metric in place of $\mathrm{F}_1$ -- comparing two
curves: $\mathrm{MCC}$-$\mathrm{F}_1$ and $\mathrm{MCC}$-$\mathrm{P}_4$. We chose the \emph{Support Vector Machine} classifier with
probabilistic output (see \cite{cortes1995support}, \cite{platt1999probabilistic}) and the Gaussian kernel.  The result -- two ROC-like
curves -- can be seen in the chart below:

\begin{center}
  \includegraphics[width=1.0\textwidth]{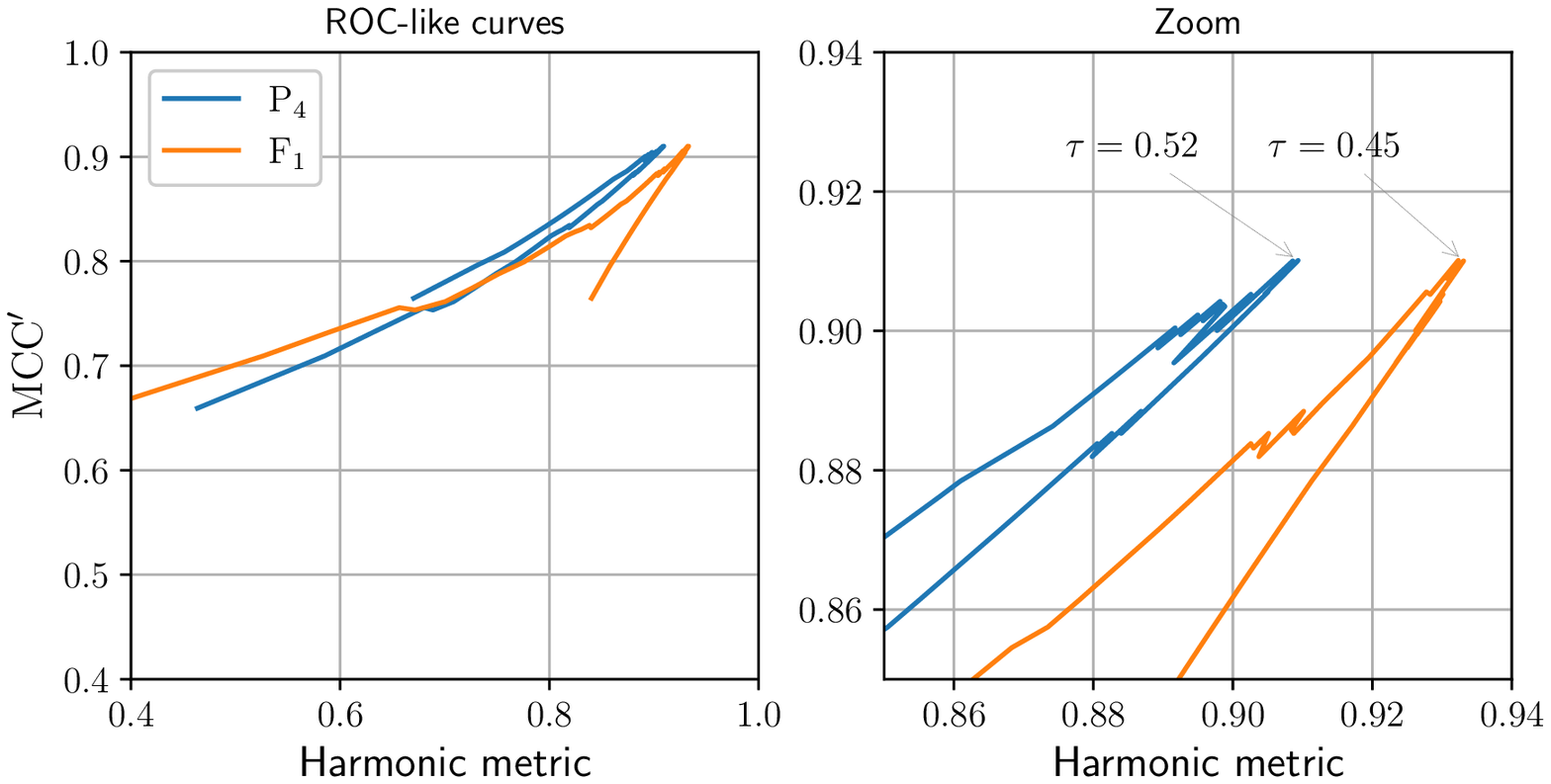}
\end{center}

The resulting graphs are similar in both cases, however for $\mathrm{P}_4$ the curve has a smaller opening angle. The critical section is captured
in the close-up on the graph to the right. The curves give different optimal threshold values: $\tau = 0.52$ for
$\mathrm{MCC}$-$\mathrm{P}_4$ curve and $\tau = 0.45$ for $\mathrm{MCC}$-$\mathrm{F}_1$. From the perspective of this single experiment --
the behavior of $\mathrm{P}_4$ metric is as expected. The difference between the two results, is since $\mathrm{P}_4$
includes two additional components comparing to $\mathrm{F}_1$ -- conditional probabilities: $P(C{-} \mid -)$ and  $P(- \mid C{-})$.

\section{Conclusions}
The definition of the new $\mathrm{P}_4$ metric presented, broadens the range of available tools for evaluating binary classifiers.  It
represents one step further in the direction indicated by $\mathrm{F}_1$. The main advantages of $\mathrm{P}_4$ are that it zeroes out when
at least one of the key four conditional probabilities also zeroes out, and that reaching a value close to $1$ requires that all four probabilities
also go to $1$.

We realize that evaluating the performance of binary classifiers is a complex problem, and we cannot expect a single metric to be the
ultimate gold standard here. Some situations may require that selected conditional probabilities be considered more significant than others.
And this, in turn, will require the development of weights like those known from $\mathrm{F}_\beta$.

The key differences between $\mathrm{P}_4$ and $\mathrm{MCC}$ are a different probabilistic interpretation and a guarantee that
$\mathrm{P}_4$ will zero out under certain conditions. Finally, their values belong to other ranges: $\mathrm{MCC} \in [-1, 1]$ and
$\mathrm{P}_4 \in [0, 1]$. The last one may be perceived as a little easier to interpret.
Despite these facts, $\mathrm{P}_4$ appears to be much closer to $\mathrm{MCC}$ than the other composite metrics.
In a situation when one uses the $\mathrm{F}_1$ however, we can frankly recommend its replacement with the $\mathrm{P}_4$.

\section{Acknowledgments}
We would like to thank the founders of \emph{SciHub} web service. Without its help, the creation of this article would not have been possible.

\appendix
\section{Symmetry} \label{symmetry}
In this appendix, we will prove the symmetry of $\mathrm{P}_4$ with respect to dataset labels swapping. By label swapping we mean renaming labels from positives to negatives and vice versa.
\begin{enumerate}
\item $\mathrm{P}_4$ is defined as the harmonic mean of $\mathrm{PREC}$, $\mathrm{REC}$, $\mathrm{SPEC}$ and $\mathrm{NPV}$.
\item Harmonic mean is a commutative operation. \label{comutative}
\item Dataset label swapping causes the following changes to the confusion matrix:
  \begin{enumerate}
  \item $\mathrm{TP}$ becomes $\mathrm{TN}$
  \item $\mathrm{TN}$ becomes $\mathrm{TP}$
  \item $\mathrm{FP}$ becomes $\mathrm{FN}$
  \item $\mathrm{FN}$ becomes $\mathrm{FP}$
  \end{enumerate}
\item After this changes to the confusion matrix: $\mathrm{PREC}$ becomes $\mathrm{NPV}$, $\mathrm{NPV}$ becomes $\mathrm{PREC}$ (see definitions in section \ref{metrics}).
\item Similarly, $\mathrm{REC}$ becomes $\mathrm{SPEC}$, $\mathrm{SPEC}$ becomes $\mathrm{REC}$ 
\item Swapping the order of the arguments of the harmonic mean does not change its value (see point \ref{comutative}) -- what ends the proof.
\end{enumerate}

\vskip 0.2in
\bibliography{extending-f1}

\end{document}